\documentclass{article}
\usepackage{spconf,amsmath,graphicx}
\usepackage{subcaption} 
\usepackage{multicol,multirow,booktabs}
\usepackage[ruled,linesnumbered,lined,boxed,commentsnumbered]{algorithm2e}
\usepackage{dblfloatfix}
\usepackage{graphicx}
\usepackage{algpseudocode}
\usepackage{threeparttable} 

\title{Enhancing Code-Switching Speech Recognition with LID-Based Collaborative Mixture of Experts Model}
\name{Hukai Huang$^{1}$, Jiayan Lin$^1$, Kaidi Wang$^1$, Yishuang Li$^3$, Wenhao Guan$^{1}$, Lin Li$^{2,3*}$, Qingyang Hong$^{1*}$
\thanks{*Corresponding authors \\ This work was supported in part by the National Natural Science Foundation of China under Grants 62276220 and 62371407.}}
\address{
  $^1$School of Informatics, Xiamen University, China \\
  $^2$School of Electronic Science and Engineering, Xiamen University, China \\
  $^3$Institute of Artificial Intelligence, Xiamen University, China
}
\begin{document}
%
\maketitle
\begin{abstract}
Due to the inherent difficulty in modeling phonetic similarities across different languages, code-switching speech recognition presents a formidable challenge. This study proposes a Collaborative-MoE, a Mixture of Experts (MoE) model that leverages a collaborative mechanism among expert groups. Initially, a preceding routing network explicitly learns Language Identification (LID) tasks and selects experts based on acquired LID weights. This process ensures robust routing information to the MoE layer, mitigating interference from diverse language domains on expert network parameter updates. The LID weights are also employed to facilitate inter-group collaboration, enabling the integration of language-specific representations. Furthermore, within each language expert group, a gating network operates unsupervised to foster collaboration on attributes beyond language. Extensive experiments demonstrate the efficacy of our approach, achieving significant performance enhancements compared to alternative methods. Importantly, our method preserves the efficient inference capabilities characteristic of MoE models without necessitating additional pre-training.
\end{abstract}
\section{Introduction}
\label{sec:intro}
Multilingual speech recognition systems are required to excel in recognizing distinct monolingual languages and, equally importantly, in effectively managing code-switching (CS) scenarios. This necessitates the system's capability to efficiently navigate the complexities of language switching within utterances, a phenomenon increasingly common in today's globalized context. Unfortunately, this issue is challenging, as similar pronunciations across various languages can easily be confused.

End-to-end speech recognition systems have gained popularity in recent years due to their simple training process and outstanding performance \cite{ctc, attention, LAS, joint,transformer-transducer,speech-transformer}. However, the end-to-end training approach usually requires a large amount of training data. Consequently, the need for more CS training data has emerged as a bottleneck, constraining recognition performance in CS scenarios. CS corpus synthesis \cite{TTS1, TTS2, TTS3} and self-supervised speech representation learning \cite{self-sup1,self-sup3,self-sup2} alleviate this data scarcity problem by synthesizing corpus or utilizing unlabeled data. Cross-language transfer learning 
 \cite{qianyi1,qianyi2,qianyi3} also effectively alleviates the data scarcity problem by utilizing monolingual corpus to improve performance in CS scenarios. From a practical point of view, mixing monolingual data from different languages with CS data to train a model that performs well in both monolingual and CS scenarios is a reasonable practice. However, the unified modeling of multilingual units may lead to the incorrect modeling of phonetically similar pronunciations across different languages. Therefore, effectively modeling diverse languages within a unified architecture presents a significant challenge 
 \cite{degrade,youxiaojianmo1,youxiaojianmo2}. 

Previous Bi-Encoder methods \cite{Biencoder-MoE, LAE, LSCA} address this issue by modeling each language separately using two parallel encoders. However, these approaches lack flexibility, exhibit high computational complexity, and struggle to support additional language extensions. Moreover, a separate pre-training initialization of each encoder on monolingual data is required to obtain better performance. The sparse Mixture of Experts (MoE) method \cite{FLR-MOE, LSR-MOE, google-smoe, MIE} aims to extract language-specific representations by introducing expert layers. Nonetheless, the expert layers in these methods struggle to achieve explicit specialized training as they do not utilize the language information inherent in the transcription text of the training data. In other words, without additional language supervision, accurate language classification is hard to achieve, which can easily disrupt the specialized training of expert networks.

This work aims to develop a bilingual model that performs well in both monolingual and CS scenarios. We propose leveraging Language Identification (LID) to guide expert routing and collaboration within the MoE model. Accurate expert routing ensures minimal interference from other language domains, effectively mitigating language confusion. Furthermore, expert collaboration based on LID weights enables comprehensive integration of language-specific representations, improving performance. Our contributions can be summarized as follows:
\begin{itemize}
    \item We show the superiority of LID-based routing in MoE models over the model without language supervision, achieving better performance and inference efficiency than bi-encoder architectures.
    \item We investigate the effectiveness of inter-group collaboration using LID weights and intra-group collaboration in an unsupervised manner, confirming their efficacy through empirical validation.
    \item Our model maintains efficiency through sparse activation while also exhibiting flexibility, allowing for adjusting the number of experts within expert groups according to the requirements of the specific task.
\end{itemize}

\section{Related Works and Motivation}
\subsection{Bi-Encoder architecture}
The Bi-Encoder architecture was first introduced in \cite{Biencoder-MoE}, where two independent encoders act as Chinese and English experts, respectively. This design necessitates pre-training each encoder on monolingual datasets to achieve specialization. Subsequently, acoustic features are processed in parallel through both encoders, yielding language-specific representations $h^{cn}$ and $h^{en}$ for Chinese and English, respectively. A linear layer at the end of the Bi-Encoder learns implicit frame-level LID by integrating $h^{cn}$ and $h^{en}$ and obtains the interpolation coefficients $\alpha_t^{cn}$ and $\alpha_t^{en}$ for each frame. Therefore, at any given frame $t$, the fusion of the two encoder outputs can be expressed as follows:
\begin{equation}
[\alpha^{cn}, \alpha^{en}]= \text{Softmax}(\text{Linear}(\text{Concat}(h^{cn}, h^{en}))
\label{eq
}
\end{equation}
\begin{equation}
h_{t}^{mix} = \alpha_t^{cn}h_t^{cn}+\alpha_t^{en}h_t^{en}
\label{eq:1_modified}
\end{equation}

LAE \cite{LAE} and LSCA \cite{LSCA} build on the Bi-Encoder architecture by introducing a language masking strategy to improve the language-specific modeling of each encoder. However, these methods struggle to overcome the inherent inefficiency of the Bi-Encoder structure since both encoders are fully involved in the forward propagation, resulting in an unavoidably large computational overhead.

\subsection{Switch Mixture of Experts}
Similar to the mechanism described in the switch-transformer \cite{switch-transformer}, the switch MoE model \cite{LSR-MOE} selects between two expert networks—one designed for Chinese and the other for English—during forward propagation at each time step. This selection is made by a linear layer within each MoE layer, acting as a gating network. The linear layer implicitly learns frame-level Language LID and decides which expert network to use based on the highest probability for each frame, as illustrated in the following equation:
\begin{equation}
r_t=\operatorname{Softmax}({W}_t h_t+{b}_t) 
\end{equation}
\begin{equation}
\boldsymbol{y}_t= \begin{cases}r_t^{c h} E_t^{c h} & \text { if }\left(r_t^{c h}>=r_t^{e n}\right) \\
r_t^{e n} {E}_t^{e n} & \text { if }\left(r_t^{c h}<r_t^{e n}\right)\end{cases}
\end{equation}
where $h_t$ denotes the input of the MoE layer, for the frame-level LID probabilities $r_t^{ch}$ and $r_t^{en}$ obtained by implicit learning, expert with higher probability is selected for forward propagation, as shown in Equation 3, ${E}_t^{ch}$ and ${E}_t^{en}$ represents the output of Chinese expert and English expert at frame $t$ respectively.
\subsection{Motivation}
The above methods exhibit inherent limitations; it prompts us to question \textit{whether a simple linear layer, without any language supervision, can learn accurate enough frame-level LID. Why do the learned weights represent language weights rather than gender or speaker-related information?} Therefore, we propose adopting a multi-task training approach for LID to explicitly endow the model with the ability to distinguish between languages. 

Furthermore, based on the original intention of MoE, we aim to integrate the knowledge from language-specific expert branches through expert collaboration. Firstly, based on language weights, a supervised LID network can determine which expert groups should participate in expert collaboration. Each expert group can have more than one expert, and the fusion features are obtained through unsupervised collaboration. This is similar to forming a committee of experts based on the task requirements during actual decision-making, followed by internal discussions within the committee to reach the final result.
\begin{figure}[]
  \includegraphics[width=\linewidth]{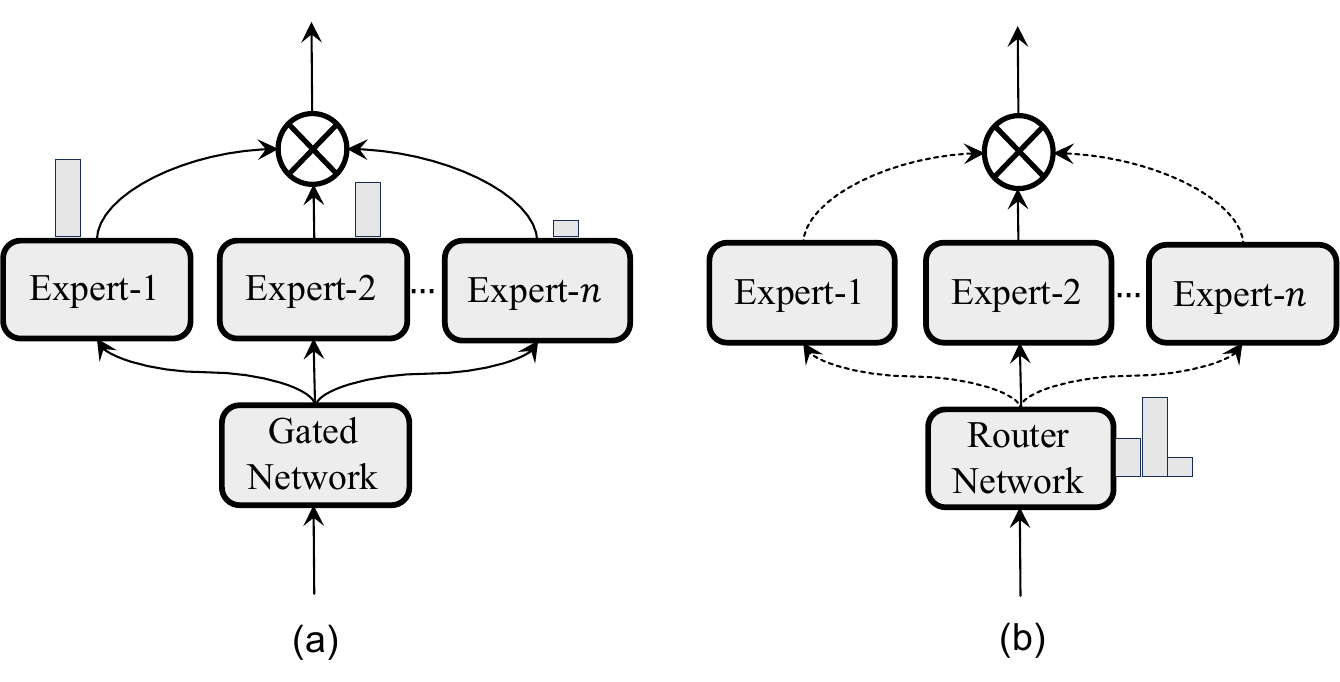} 
  \caption{Two types of MoE structures. (a) Dense-MoE, all experts will be activated. (b) Sparse-MoE, only some experts are activated.}
  \label{fig:example1}
\end{figure}

\section{Proposed Method}
\subsection{Dense-MoE \&\& Sparse-MoE}
Fig.\ref{fig:example1} illustrates two types of MoE models. Subfigure (a) depicts Dense-MoE, which was the original form of MoE when it was first proposed \cite{origin-moe}. In Dense-MoE, all expert networks participate in forward propagation and are weighted and fused according to the weights obtained from gating networks to obtain more comprehensive features. The previously mentioned Bi-Encoder architecture \cite{Biencoder-MoE} can be considered a variant of Dense-MoE because all expert networks are involved in the computation. Subfigure (b) represents Sparse-MoE models, where only a tiny portion of expert networks are typically selected to participate in computation. Sparse-MoE has recently gained popularity in large language models because it can increase the model parameter size while maintaining constant inference efficiency\cite{switch-transformer}. 
\subsection{LID-driven expert routing mechanism}
\begin{figure}[]
  \includegraphics[width=\linewidth]{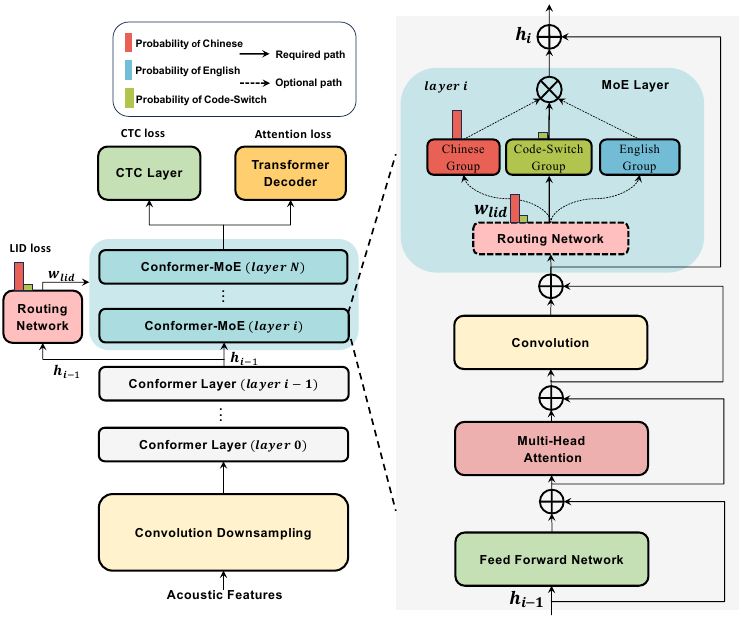} 
  \caption{The architecture of Collaborative-MoE. A routing network provides route guidance to subsequent MoE layers, specializing in the expert network and providing collaborative weights (as shown by the color bar in the figure).}
  \label{fig:example2}
\end{figure}

Our proposed model combines the advantages of Dense-MoE and Sparse-MoE for modeling Chinese and English. As shown in Fig.\ref{fig:example2}, the proposed model consists of two main parts: the first $i$ layers consist of the vanilla Conformer \cite{Conformer} layers, while the last several layers replace the FFN at the end of the Conformer with MoE layers, forming the Conformer-MoE layers. The bottleneck layer features obtained from the first $i$ Conformer layers already contain rich language information and are fed into a routing network for LID. To adapt to the CS task, we treat CS utterances as a separate category, which requires the model to detect language switches in the input and distinguish them from monolingual utterances. This adaptation facilitates the model in extracting more language-specific features. The routing network, comprising a linear layer followed by a global pooling layer, distributes features to expert groups for processing. For bottleneck layer features $h_{i-1}$, the modeling process is represented as follows:
\begin{equation}
h_{\text{lid}} = \text{Linear}(\text{GlobalPooling}(h_{i-1}))
\label{eq:1}
\end{equation}
\begin{equation}
\boldsymbol{P} = \frac{\exp\left(h_{\text{lid}}^j / T\right)}{\sum_{j=1}^3 \exp\left(h_{\text{lid}}^j / T\right)}
\end{equation}
where $h_{\text{lid}}$ represents the logits for LID, which are further used to obtain $\boldsymbol{P}$ based on Equation 6. $\boldsymbol{P}=[p_{\text{ch}}, p_{\text{en}}, p_{\text{cs}}]$ represents LID classification probability, the routing network will determine whether to activate the Chinese or English expert group based on the larger value between $p_{\text{ch}}$ and $ p_{\text{en}}$. This selection will be applied to all subsequent Conformer-MoE layers. By adopting this approach, the updates of Chinese expert parameters will not be interfered with by English data, effectively mitigating language confusion and achieving specialized training. The CS group in the MoE layer will model all language data, and there can be more than one expert within the expert group. For example, the CS group can be composed like Dense-MoE, as shown in Fig. 1(a).

\subsection{Expert collaboration mechanism}

We employed two types of expert collaboration mechanisms: one based on LID-weighted inter-group expert collaboration and the other being intra-group expert collaboration. For inter-group expert collaboration, in Equation 6, we introduced a temperature coefficient T to smooth the LID output probability $\boldsymbol{P}=[p_{\text{ch}}, p_{\text{en}}, p_{\text{cs}}]$. This is primarily to avoid sharp distributions leading to one expert's weight almost reaching $1$. After removing the probabilities of expert groups not selected in $\boldsymbol{P}$ and re-normalizing, we obtained expert collaboration weights $w_{lid}=[w_{op},w_{cs}]$ and performed inter-group representation fusion. The term $w_{op}$ represents the normalization of the larger value between $p_{\text{ch}}$ and $p_{\text{en}}$. As for intra-group expert collaboration, since the language has already been determined, we adopted an unsupervised gating network similar to Dense-MoE to model differences outside of language and obtain better representations.

\subsection{Loss function}
For the routing network, we use cross-entropy loss to train the LID task, denoted as follows:
\begin{equation}
\mathcal{L}_{LID} = \text{CrossEntropy}(h_{lid}, y_{lid})
\end{equation}
where $y_{lid}$ is the language ID of utterance. The total loss is expressed as follows:
\begin{equation}
\mathcal{L}=\lambda_{asr} \mathcal{L}_{att}+\left(1-\lambda_{asr}\right) \mathcal{L}_{ctc}+\lambda_{lid} \mathcal{L}_{LID}
\end{equation}
where $\lambda_{asr}$ and $\lambda_{lid}$ are hyperparameters that control the weights of the ASR and LID tasks, respectively. The utterance-level LID task is trained jointly with the ASR task due to its relative simplicity without extra pre-training.
\section{Experiments}
\subsection{Dataset}
Our experiments are conducted on the ASRU-2019 Mandarin-English code-switching challenge dataset \cite{asru}, and a subset of 460 hours of Librispeech \cite{librispeech} was also included for training. The details of the dataset are shown in Table 1.
\begin{table}[!h]
\caption{Details of the dataset used for the experiments, corpus ‘ALL’ means mixing all datasets for training. We use ‘CN,’ ‘EN,’ and ‘CS’ to denote Mandarin, English, and Code-switching, respectively.}
\setlength{\tabcolsep}{6pt} 
\setlength{\intextsep}{10pt plus 2pt minus 2pt}
\renewcommand{\arraystretch}{1.2} 
\begin{tabular}{ccccccll}
\cline{1-6}
\multirow{2}{*}{Language} & \multirow{2}{*}{Corpus} & \multicolumn{2}{c}{Duration(h)}             & \multicolumn{2}{c}{Utterance(k)}            &  &  \\ \cline{3-6}&  & Train & Test  & Train   & Test  &  &  \\ \cline{1-6}
CN  & ASRU-Man  & 485.8  & 12.4  & 545.8  & 14.0 && \\
EN  & Librispeech  & 464.2 & 5.4  & 132.5  & 2.6 &&  \\
CS & ASRU-CS  & 197.8 & 20.3 & 184.5 & 16.2 &&  \\
Mix & ALL & 1147.8  & 38.1 & 862.8 & 32.8  &&  \\ \cline{1-6}

\end{tabular}
\end{table}

\subsection{Model configuration}
We conduct our experiments on the WeNet \cite{wenet} toolkit. For acoustic features, 80-dimensional log-mel filter banks (FBank) are extracted with a step size of 10ms and a window size of 25ms. SpecAugment \cite{specaugment} is applied during all training stages. 

All of our experiments use a 6-layer transformer as the decoder, and the encoders are all stacked with Conformer \cite{Conformer}. The encoders' configurations used for the experiments in this paper are as follows: \textbf{Baseline}: 12-layer Conformer. \textbf{Bi-Encoder} \cite{Biencoder-MoE, LSCA}: two parallel encoders, each consisting of 12 layers of Conformer. \textbf{LAE} \cite{LAE}: a 6-layer shared module and a 6-layer language-specific module for each language. \textbf{Sparse MoE} \cite{LSR-MOE}: 12-layer Conformer-MoE, each expert layer contains two expert networks and a routing network. \textbf{Proposed Model}: 6-layer Conformer shared module and a 6-layer Conformer-MoE module, each expert layer consists of three expert groups. Additionally, we designed a \textbf{Dense-MoE} model for comparison with our proposed model. The critical difference in Dense-MoE is that the expert layers in the last six Conformer-MoE layers use the structure shown in Fig.\ref{fig:example1}(a) without the inclusion of additional LID information. The expert layers contain four language-agnostic expert blocks, which conduct intra-group collaboration based on a gating network. Correspondingly, we expanded the FFN dimensions in the Conformer Baseline at the corresponding positions four times to create a \textbf{Conformer Large} with comparable parameters. 

The attention dimension, number of attention heads, feed-forward dimension, and convolutional kernel size are fixed to 256, 4, 2048, and 31, respectively. We set the $\lambda_{asr}=0.7$, $T=10$ in the training stage of the proposed model. We will scale the LID loss to align with the magnitude of the ASR loss and then apply a coefficient of $\lambda_{lid}=0.1$, and the Adam optimizer is used with a learning rate schedule with 25,000 warm-up steps. We use character error rate (CER), word error rate (WER), and mix error rate (MER) to evaluate the performance of Mandarin, English, and CS scenarios, respectively.

\subsection{Results of proposed model and baselines}
\begin{table*}[h]
\centering  
\caption{Performance of different systems on the test sets. The real-time factors (RTF) of the different systems tested on a CPU are also shown in this table. The model's computational complexity is assessed by calculating the floating-point operations (FLOPs) for a 20-second input sample. The results of the three test sets and the average error rate of the three test sets are also shown in the table.}
\label{tab:example2}
\setlength{\tabcolsep}{7pt} 
\renewcommand{\arraystretch}{1.2} 
\scalebox{0.9}{
\begin{tabular}{ccccccccc}
\toprule[\heavyrulewidth]
\multicolumn{1}{c}{\multirow{2}{*}{System}} & \multicolumn{1}{c}{\multirow{2}{*}{Model}} & \multirow{2}{*}{Params} & \multirow{2}{*}{FLOPs $\downarrow$} & RTF $\downarrow$  & CN & EN & CS & \multirow{2}{*}{Average} \\
\multicolumn{1}{c}{} & \multicolumn{1}{c}{} &     &       &  ($10^{-2}$) & CER(\%) & WER(\%) & MER(\%) & 
\\
\midrule
$S1$ & Conformer Baseline & 51M & 25.6G & 4.21   & 3.35 & 6.89 & 9.93 & 7.18 \\
$S2$ & Conformer Large & 72M & 36.8G & 4.75 & 3.15 & 6.49 & 9.74 & 6.95 \\
$S3$ & Bi-Encoder MoE & 85M & 48.4G & 5.89  & 3.01 & 5.94 & 9.67 & 6.79 \\
$S4$ & Bi-Encoder LSCA & 88M & 51.2G & 5.79  & 2.82 & 5.72 & 9.52 & 6.62 \\
$S5$ & LAE Conformer & 75M & 38.4G & 4.80 & 2.98  & 5.86 & 9.65 & 6.76 \\
$S6$ & Sparse-MoE & 63M & 26.5G & 4.28  & 3.13 & 6.56 & 9.78 & 6.97 \\
\midrule 
$S7$ & Dense-MoE (ours) & 72M & 37.5G & 4.85 & 2.85  & 5.96 & 9.45 & 6.62 \\
$S8$ & Collaborative-MoE (ours) & 72M & 32.1G & \textbf{4.52} & \textbf{2.59} & \textbf{5.40}  & \textbf{9.33} & \textbf{6.45} \\ \bottomrule[\heavyrulewidth]
\end{tabular}
}
\begin{tablenotes}
    \item[1] * Note: The above results are achieved with our experimental configuration on the training set ‘ALL’.
\end{tablenotes}
\end{table*}

As shown in Table \ref{tab:example2}, our proposed model $S8$ outperforms previous bi-encoder methods $S3$, $S4$, and $S5$ while significantly reducing GFLOPs and RTF. This improvement is mainly due to its compact topology, resembling a single encoder, and benefiting from the efficient inference of sparse MoE. Compared to the $S1$, we achieve an average relative performance improvement of 10.17\% across the three test sets, with relative performance improvements of 22.7\%, 21.6\%, and 6\% on CN, EN, and CS, respectively, while maintaining comparable computational costs.

The combined results of $S1$, $S2$, and $S7$ suggest that merely increasing FFN dimensions has limited effectiveness. In $S7$, the Dense-MoE with four expert networks in the MoE layer outperforms the Conformer Large model $S2$ with equivalent parameters due to the unsupervised learning of gating network weights. This highlights the importance of unsupervised collaboration in Dense-MoE. No additional LID loss was applied in Dense-MoE so that the gating network might have learned some language preferences or modeling differences beyond languages in an unsupervised manner.

Both systems, S8 and S7, have four expert networks, but S8 introduces explicit LID, making the expert networks language-specific. Within the CS group, two experts are designed to collaborate with the group based on the Dense-MoE manner. Each Chinese and English expert group has one expert network, which collaborates with the CS group based on LID weights. We observe a more significant performance improvement of the $S8$ compared to $S7$ in CN and EN scenarios. This is primarily due to the explicit LID distributing data of different languages to specific expert groups, effectively mitigating language confusion caused by mixed-language training. This allows the expert groups to develop language-specific solid modeling capabilities. Additionally, due to the proposed model's sparse MoE characteristics, it has lower GFLOPs and RTF compared to S2 and S7 with equivalent parameters. We note that the performance improvement of Sparse-MoE $S6$ is relatively limited, resembling that of in switch MoE \cite{LSR-MOE}. This is believed to be due to the lack of information on language supervision, making expert selection more challenging and less interpretable. In summary, the results indicate that incorporating LID information is necessary and effective in improving performance.

\subsection{Ablation study}

\begin{table}[h]
\centering
\caption{The impact of different expert network distributions on model performance, where (CN,EN, CS) represent the number of experts within the Chinese, English, and code-switching groups, respectively.}
\label{tab:example1}
\setlength{\tabcolsep}{5pt} 
\renewcommand{\arraystretch}{1.0} 
\scalebox{0.9}{
\begin{tabular}{ccccccc}
\toprule[\heavyrulewidth]
\multirow{2}{*}{(CN,EN,CS)}   & CN & EN & CS & LID \\
       & CER(\%) & WER(\%) & MER(\%) & Acc(\%) \\ \midrule
(2,1,1) & 2.49   & 5.42 & 9.42  & 99.4 \\ 
(1,2,1) & 2.65   & 5.31  & 9.46 & 99.4\\
(2,2,0) & 2.67   & 5.45 & 9.55  & 99.4\\
(1,1,2) & 2.59   & 5.40  & 9.33 & 99.4\\

\bottomrule[\heavyrulewidth]
\end{tabular}
}
\end{table}
In this subsection, we investigate how the distribution of expert groups affects the performance across various language test sets. As mentioned earlier, owing to the LID-based routing mechanism, the number of experts within each language-specific group can be flexibly adjusted, and they work in an intra-group collaborative manner. As illustrated in Table 3, setting the number of experts within a specific language expert group to 2 enhances its performance for the corresponding language. However, given that the CS group encompasses all languages, leveraging intra-group collaboration within it yields the best overall results. The average utterance-level LID accuracy on the three test sets provided in Table 3 demonstrates the reliability of supervised routing.

We also experimented with removing the CS group, forcing the model to choose between the Chinese and English groups. The results show a significant performance degradation in CS scenarios and no advantage in CN and EN scenarios. In this case, the LID weights did not function effectively, and the lack of inter-group collaboration could cause this impact. These experiments also demonstrate that if we want the model to focus on a particular language, we can allocate more experts to the expert group of that language. This flexible design approach is challenging for other sparse MoEs that are not based on LID.

\section{Conclusions}
This work proposes Collaborative-MoE, a MoE model based on LID weight routing and collaboration. The routing network trained based on the LID task benefits from introducing language information, enabling accurate expert routing, and ensuring specialized training and inference for language-specific expert groups. A series of experiments demonstrate the effectiveness of the proposed routing based on LID weights and inter-group collaboration while also exploring and validating the effectiveness of intra-group collaboration. Experimental results indicate that the proposed approach achieves significant performance improvements compared to baselines and other methods while maintaining efficient inference. Moreover, our method features a simple training process that does not require additional pre-training, allowing for training from scratch with mixed-language data. In the future, we will explore better ways to integrate language information into MoE and experiment with more languages and larger models.


\bibliographystyle{IEEEbib}
\bibliography{refs}

\end{document}